\documentclass[runningheads]{llncs}
\usepackage[nolist]{acronym}
\usepackage{subfig}
\usepackage{amsmath}
\usepackage{amssymb}
\usepackage{commath}
\usepackage{booktabs}
\usepackage{multirow}
\usepackage{hyphenat}
\usepackage{graphicx}
\usepackage{comment}                

\usepackage[table, dvipsnames]{xcolor}
\definecolor{vinaygray}{gray}{0.9}

\newcommand{\lbleq}[1]{\label{eq:#1}}

\newcommand{\repeatthanks}{\textsuperscript{\thefootnote}}

\begin{document}
%
\title{Generating Synthetic Handwritten Historical Documents With OCR Constrained GANs}

    \author{Lars~V\"ogtlin\thanks{Both authors contributed equally to this work.} \and
    Manuel~Drazyk\repeatthanks \and
    Vinaychandran~Pondenkandath \and
    Michele~Alberti \and
    Rolf~Ingold}
    
    \authorrunning{V\"ogtlin et al.}

    \institute{\textit{Document Image and Voice Analysis Group (DIVA)} \\
    University of Fribourg, Switzerland \\
    \email{\{firstname.lastname\}@unifr.ch}}

%
\maketitle              
\begin{acronym}[UMLX]
 \acro{DIA}{Document Image Analysis}
 \acro{OCR}{Optical Character Recognition}
 \acro{NST}{Neural Style Transfer}
 \acro{GAN}{Generative Adversarial Network}
 \acro{cGAN}{conditional \acf{GAN}}
 \acro{CNN}{Convolutional Neural Network}
 \acro{RNN}{Recurrent Neural Network}
 \acro{CRNN}{Convolutional Recurrent Neural Network}
 \acro{ReLU}{Rectified Linear Unit}
 \acro{LSTM}{Long Short-Term Memory}
 \acro{GT}{Ground Truth}
 \acro{CTC}{Connectionist Temporal Classification}
 \acro{HTR}{Handwritten Text Recognition}
 \acro{TR}{Text Recognizer}
 \acro{CER}{Character Error Rate}
 \acro{WER}{Word Error Rate}
 \acro{SHI}{Synthetic Historical Image}
 \acro{RHI}{Real Historical Image}
 \acro{STI}{Synthetic Template Image}
 \acro{SD}{Style Discriminator}
 \acro{RD}{Reading Discriminator}
 \acro{CycleGAN-RD}{CycleGAN with an added \ac{RD}}
 \acro{CycleGAN-2RD}{CycleGAN with two \acp{RD}}
 \acro{PreGEN}{pretraining on the template documents}
 \acro{PreIAM}{pretraining on the IAM dataset}
 \acro{NoPre}{without pretraining}
 \acro{N}{with Gaussian noise}
 \acro{Reg}{with regeneration of the template after every epoch}
 \acro{C-RD}{\ac{RD} trained with the CycleGAN-RD}
 \acro{EF-RD}{\ac{RD} trained with the Evaluation Framework}
 \acro{VAE}{Variational Autoencoder}
 \acro{HBA}{Historical Book Analysis Competition}
\end{acronym}

%
%

\begin{abstract}
We present a framework to generate synthetic historical documents with precise ground truth using nothing more than a collection of unlabeled historical images. 
Obtaining large labeled datasets is often the limiting factor to effectively use supervised deep learning methods for \ac{DIA}. 
Prior approaches towards synthetic data generation either require human expertise or result in poor accuracy in the synthetic documents. 
To achieve high precision transformations without requiring expertise, we tackle the problem in two steps.
First, we create template documents with user-specified content and structure.
Second, we transfer the style of a collection of unlabeled historical images to these template documents while preserving their text and layout. 
We evaluate the use of our synthetic historical documents in a pre-training setting and find that we outperform the baselines (randomly initialized and pre-trained). 
Additionally, with visual examples, we demonstrate a high-quality synthesis that makes it possible to generate large labeled historical document datasets with precise ground truth.

\keywords{OCR \and CycleGAN \and synthetic data \and historical documents.}
\end{abstract}

\section{Introduction}

\begin{figure}[t]
    \centering
\subfloat[]{{%
  \includegraphics[width=0.3\textwidth]{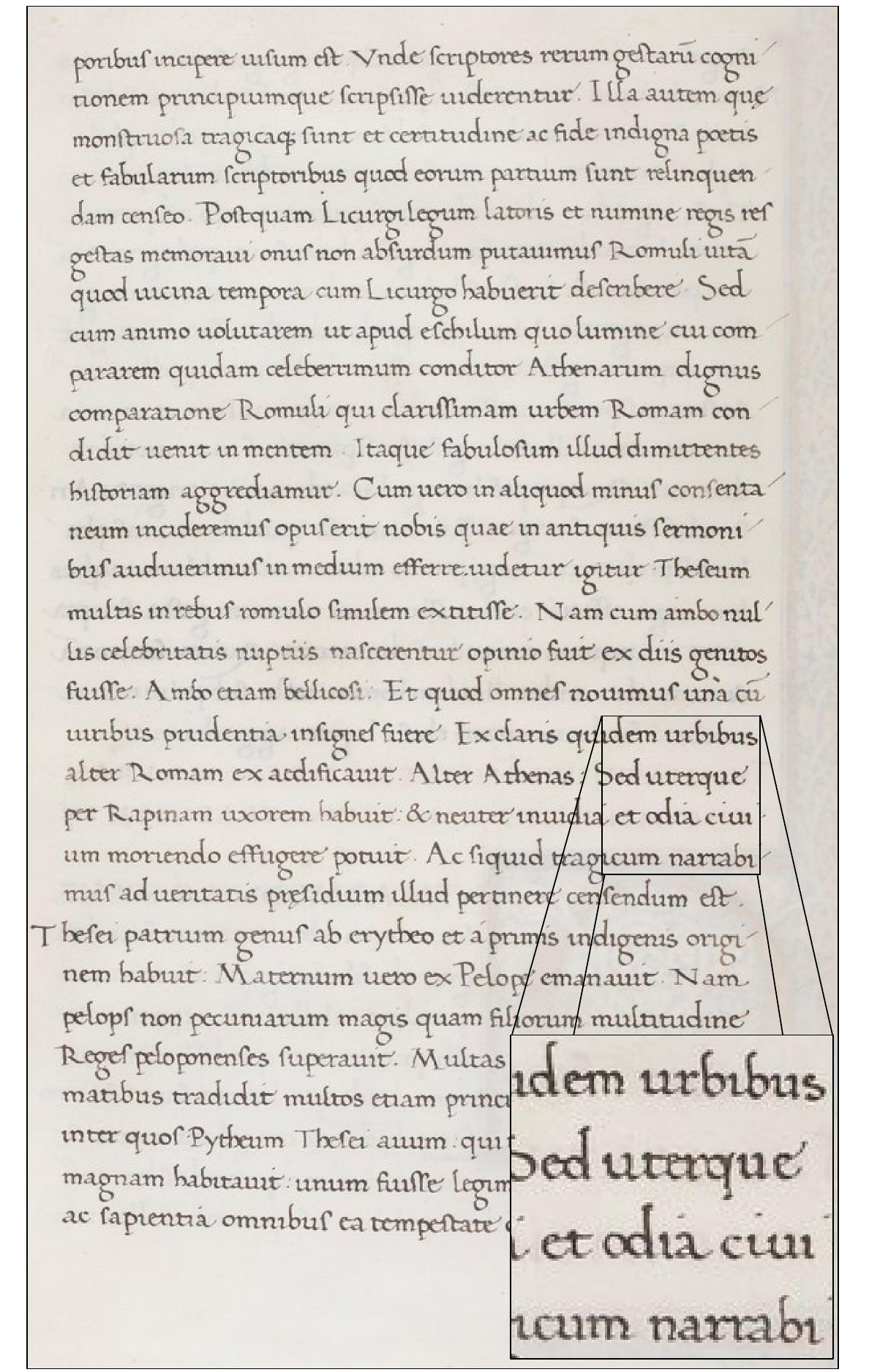}}%
  \label{fig:intro_real}}%
  \hspace{0.4em}
  \subfloat[]{{%
  \includegraphics[width=0.3\textwidth]{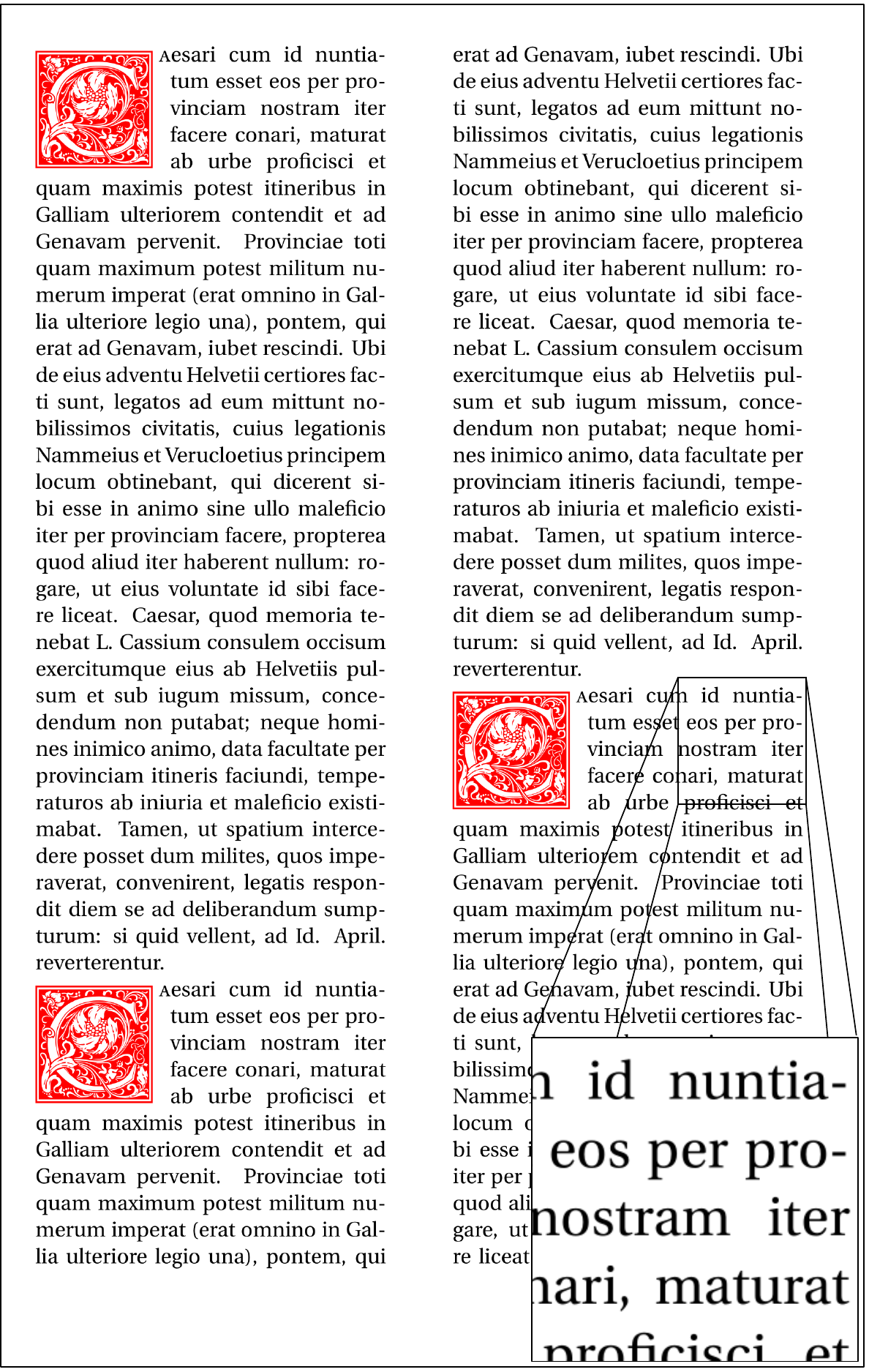}}%
  \label{fig:intro_template}}%
  \hspace{0.4em}
  \subfloat[]{{%
  \includegraphics[width=0.3\textwidth]{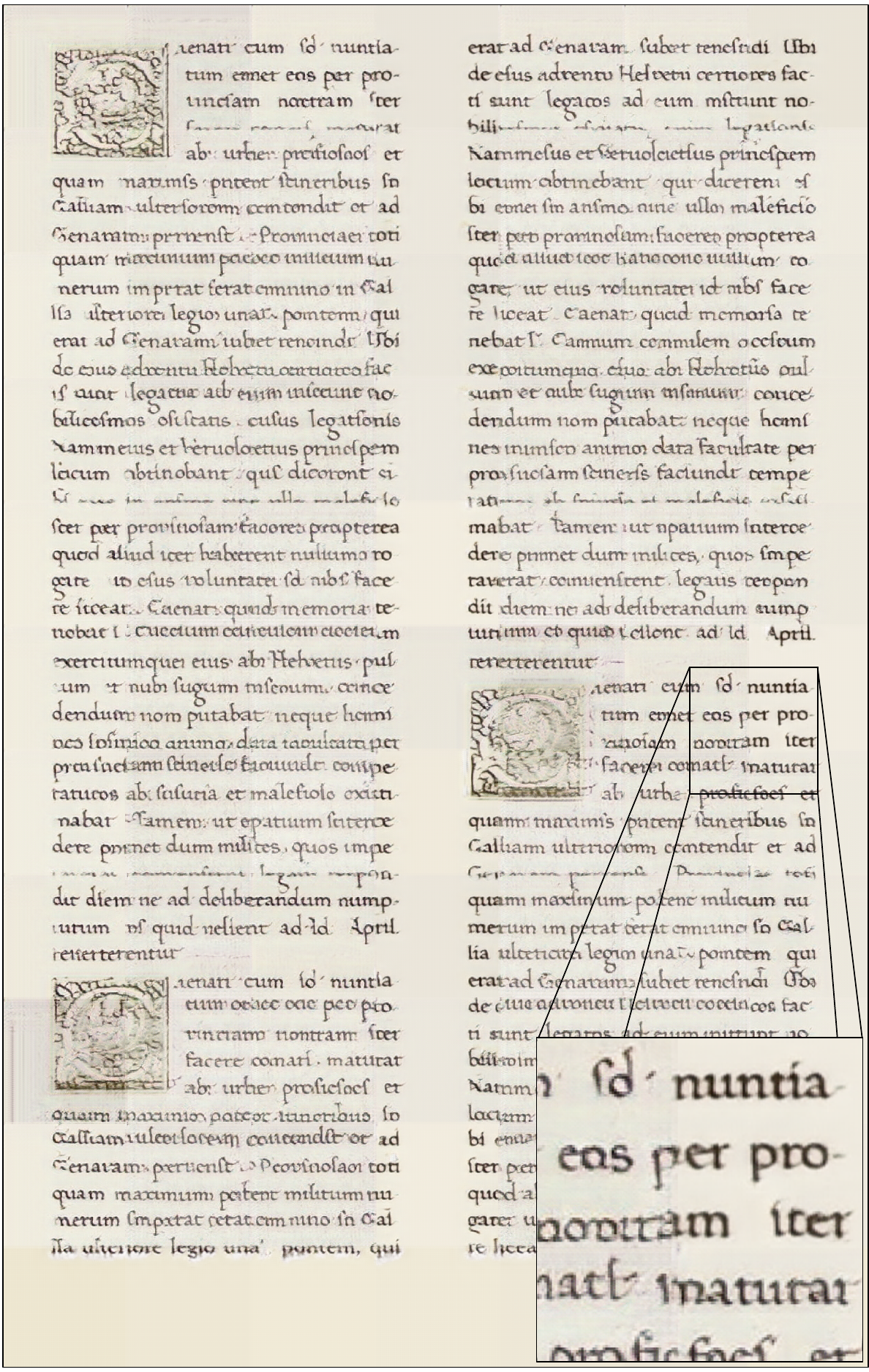}}%
  \label{fig:intro_synthetic}}%
    \caption{
    Inputs for the second step of our framework and the output of the network. 
    (a) represents the style template for our output document. 
    (b) shows a source document that was generated using \LaTeX. 
    (c) shows the corresponding transformed version of the template image (b). 
    The transformation between (b) and (c) preserves overall structure and content.
    }
    \label{fig:intro_fig}
\end{figure}

Large labeled datasets play a major role in the significant performance increases seen in \ac{DIA} and computer vision over the last decade.
These datasets -- often containing millions of labeled samples -- are typically used to train deep neural networks in a supervised setting, achieving state-of-the-art performance in tasks such as text line segmentation~\cite{alberti2019labeling}, \ac{OCR}~\cite{blucheScanAttendRead2017} or layout analysis~\cite{scius-bertrandLayoutAnalysisText2019}.
However, such methods are much more challenging to train in settings where no labeled data is available. 
The size of labeled datasets is limited to a few hundred or thousand samples -- as is often the case with historical documents~\cite{journetDocCreatorNewSoftware2017a,clausnerAletheiaAdvancedDocument2011}.

Common strategies to deal with limited labeled data include (1) transfer-learning, (2) synthesizing artificial data, or (3) unsupervised learning.
In (1) typical procedure is to train a deep neural network on similar data and then fine-tune this network on the small labeled target dataset.
The success depends on having datasets similar enough to the target dataset to perform pre-training.
(2) has been an active area of \ac{DIA} research. 
Baird~\cite{bairdDocumentImageDefect1992}, Kieu et
al.~\cite{kieuCharacterDegradationModel2012} and Seuret et
al~\cite{seuretGradientdomainDegradationsImproving2015a} focus on degrading real
document images using defect models to augment datasets.
Other tools such as DocEmul~\cite{capobiancoDocEmulToolkitGenerate2017a} and
DocCreator~\cite{journetDocCreatorNewSoftware2017a} aim to create synthetic
document images using a combination of user-specified structure, background
extraction, degradation methods, and other data augmentation approach.  
However, such approaches still require human expertise in designing appropriate
pipelines to generate realistic documents.
When large unlabeled datasets are available for the target task, a common practice is to use unsupervised learning methods such as autoencoders~\cite{masciStackedConvolutionalAutoEncoders2011} to learn representations.
However, recent work~\cite{albertiPitfallUnsupervisedPreTraining2017} shows that autoencoders trained for reconstruction are not useful for this task.
Another possibility is to use unlabeled data in a Generative Adversarial~\cite{goodfellowGenerativeAdversarialNetworks2014,zhuUnpairedImagetoimageTranslation2017} setting to synthesize artificial data that looks similar in appearance to the unlabeled data.

%
More recent work in document image synthesis has used deep learning, and \ac{GAN} based approaches.
But these approaches\newline~\cite{pondenkandathHistoricalDocumentSynthesis2019a,tensmeyerGeneratingRealisticBinarization2019,guanImprovingHandwrittenOCR2020,kangGANwritingContentConditionedGeneration2020} result in various issues: the produced data matches the overall visual style of historical documents but fail to produce meaningful textual content; they require paired datasets, which defeats the purpose of using unlabeled data; only create text of fixed length.

In this paper, we present a framework to generate historical documents without relying on human expertise or labeled data\footnote{https://github.com/DIVA-DIA/Generating-Synthetic-Handwritten-Historical-Documents}. 
We approach this problem in two steps.
First, we create template document images that contain user-specified content and structure using LaTeX\footnote{This can be done with any other word processing tool such as MS Word.}.
Second, using the user-specified template documents and a collection of unlabeled historical documents, we learn a mapping function to transform a given template document into the historical style while preserving the textual content and structure present in the template document.



We evaluate the usefulness of our synthetically generated images by measuring the performances of a deep learning model on the downstream task of \ac{OCR}. 
Specifically, we measure the performances of this model when (1) trained only on the target dataset (St. Gall \cite{fischer2011TranscriptionAlignmentLatin}), (2) pre-trained on a similar dataset (IAM Handwriting database \cite{marti2002iam}) and then fine-tuned on the target dataset and finally when (3) pre-trained on our synthetic images and then fine-tuned on the target dataset. 
This will allow us to compare against a standard supervised baseline as well as a reasonable transfer learning baseline.
Our empirical experiments show that, the model pre-trained on our synthetic images (see point 3 above) is outperforming the supervised and transfer learning baselines by 38\% and, respectively, 14\% lower \acf{CER}. 


\subsection*{Main Contribution}
This paper extends the existing work on synthetic document generation by providing a general framework that can produce realistic-looking historical documents with a specific style and textual content/structure. 
We introduce a two-step CycleGAN based process that leverages two \ac{TR} networks to condition the learning process. 
This additional signal let us overcome the main limitations of previous work and enable us to obtain significantly better performance measured on a robust set of benchmarks.

\begin{figure}[t]
\begin{minipage}{0.65\textwidth}
\centering
\subfloat[HBA]{{%
  \includegraphics[width=0.45\textwidth]{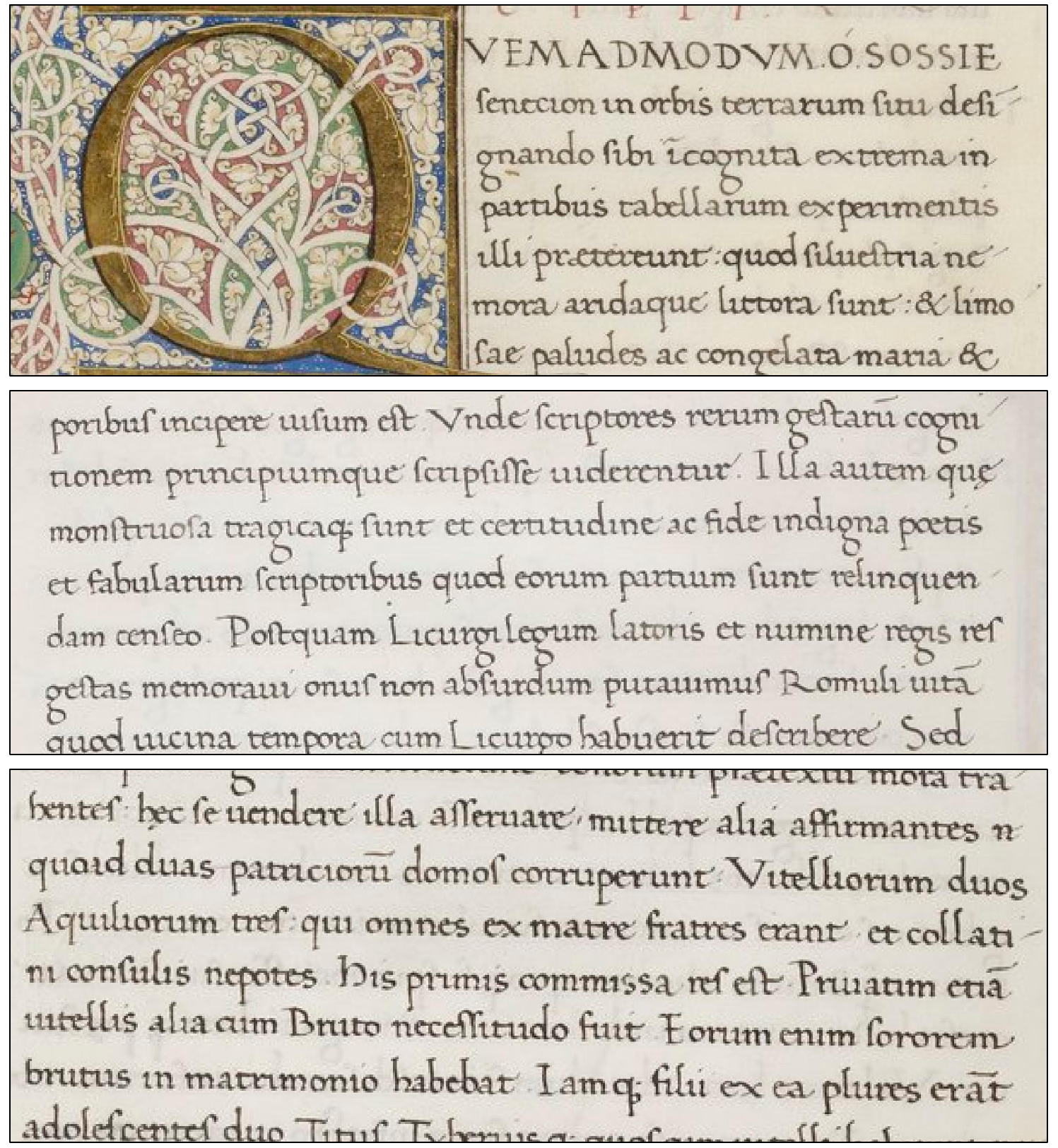}}%
  \label{fig:dataset_hba}}%
  \quad
\subfloat[StGall]{{%
  \includegraphics[width=0.45\textwidth]{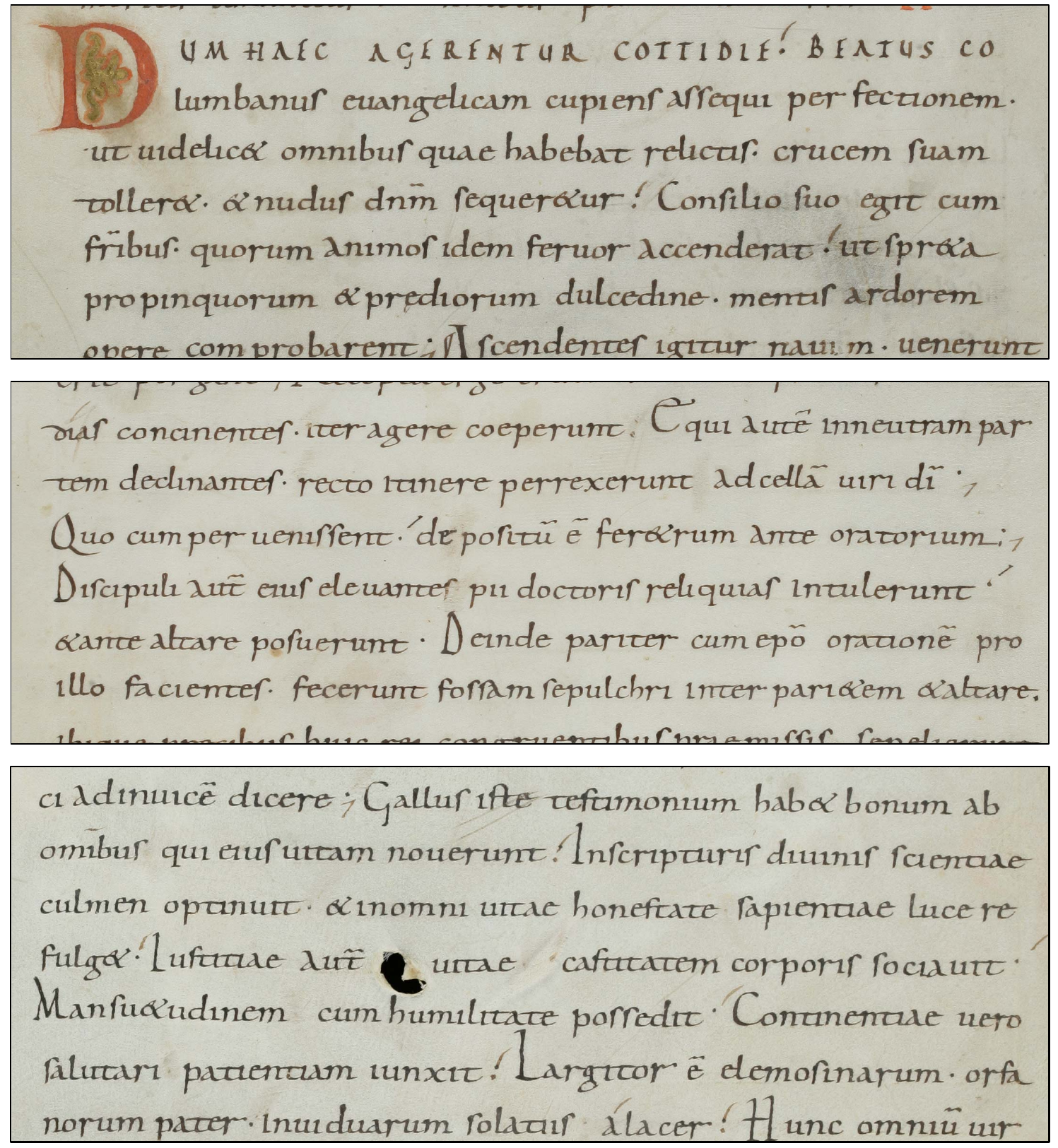}}%
  \label{fig:dataset_sg}}%
  \quad
\subfloat[IAM]{{%
  \includegraphics[width=0.45\textwidth]{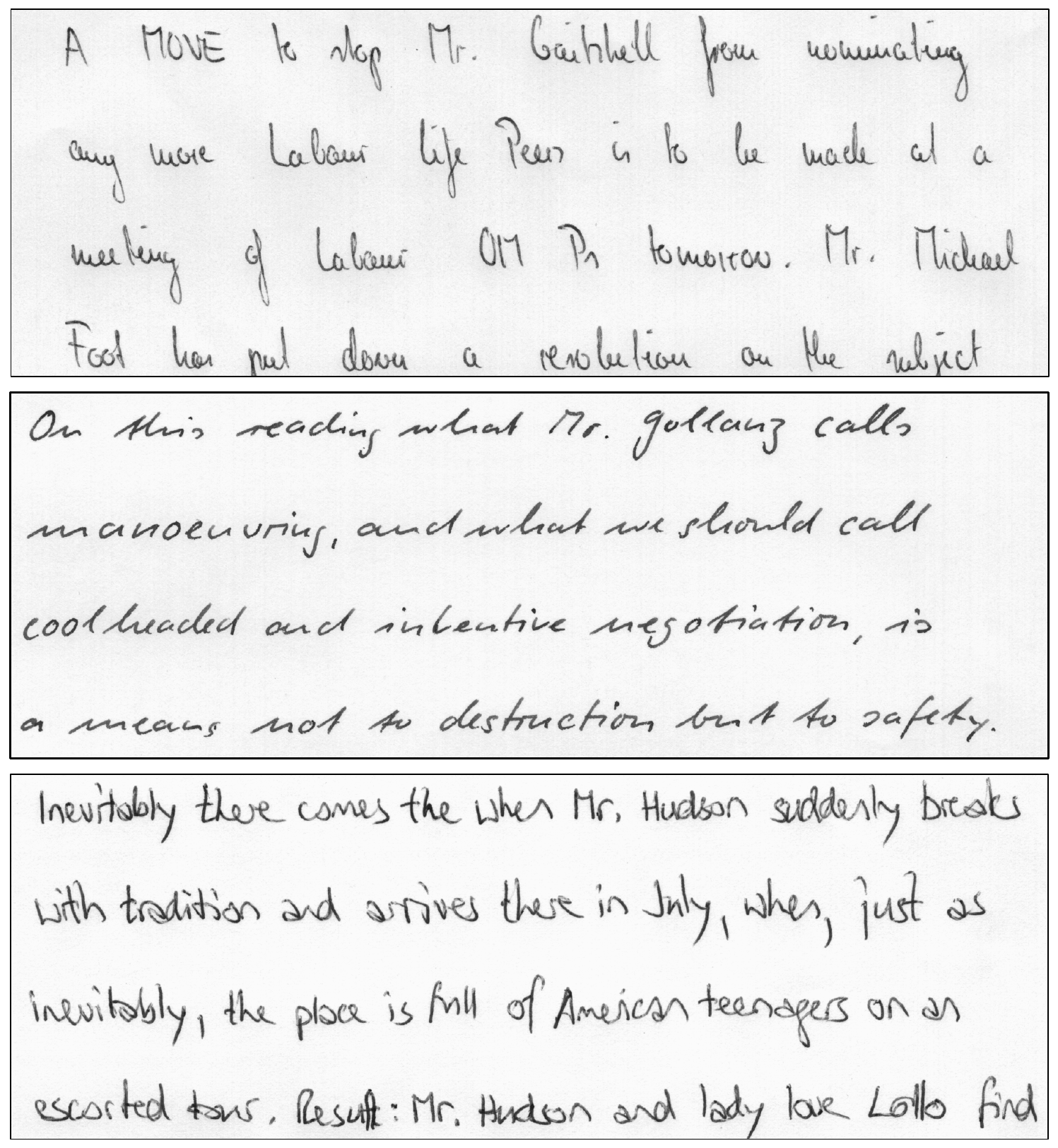}}%
  \label{fig:dataset_iam}}%
\caption{
We use the HBA dataset (a) as the target historical style, the Saint Gall dataset (b) for evaluating our synthetic data in a pre-training setting and the IAM Handwriting Database (c) as a pre-trained baseline.
}
\label{fig:dataset_all}
\end{minipage}\hfill
\begin{minipage}{0.3075\textwidth}
    \vspace{-0.5cm}
    \center{\includegraphics[width=\textwidth]
    {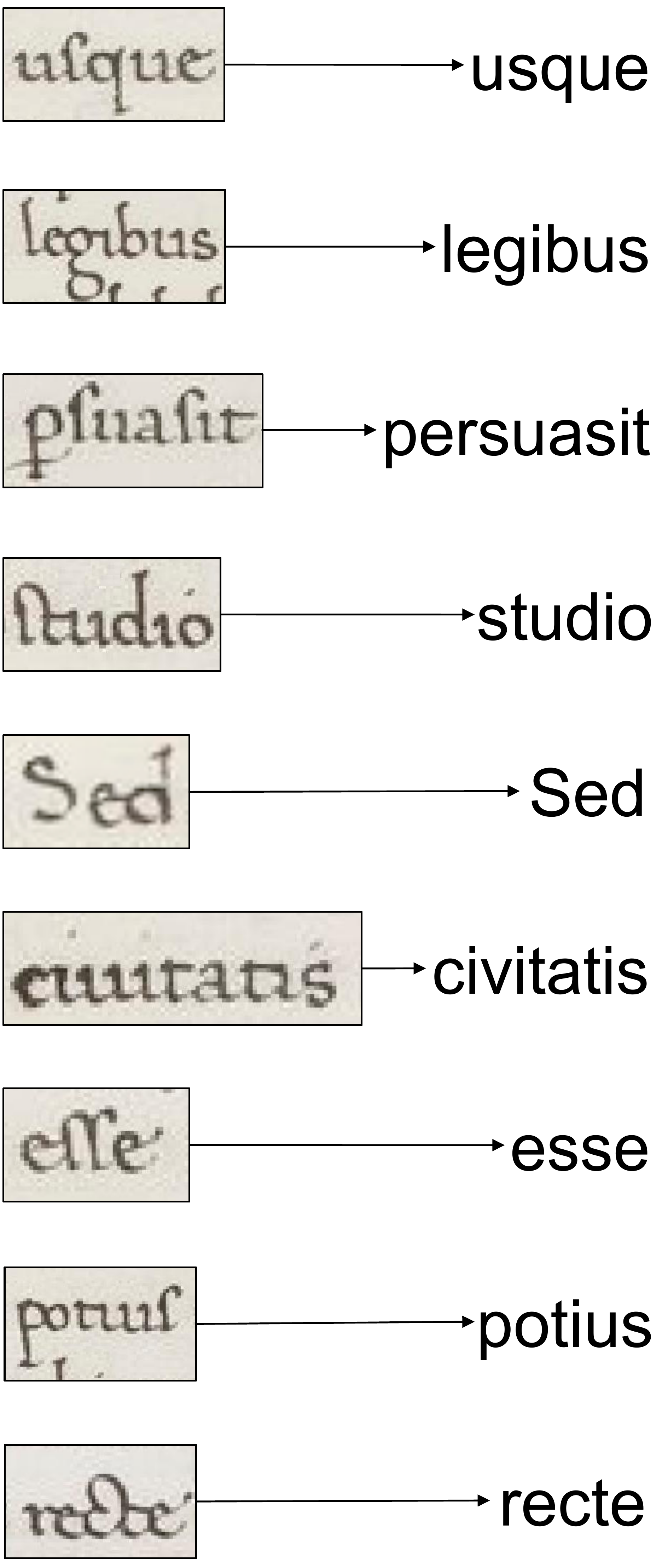}}
    \caption{Samples from the hand annotated subset of the HBA dataset used for validation purposes.}
    \label{fig:hba_words}
\end{minipage}
\end{figure}

\section{Datasets}\label{sec:datasets}

In this work, we use three datasets: the user-specified template document dataset (source domain dataset; see Section~\ref{sec:source_dataset}); a dataset of real unlabeled historical documents (target domain dataset) whose style which we want to learn in the transformation function; a dataset of real labeled historical documents (evaluation dataset) with transcription ground truth that we use to evaluate our methods.  

\subsection{Source Domain Dataset}\label{sec:source_dataset}
We create a collection of template documents with user-specified content and structure as Pondenkandath et al.~\cite{pondenkandathHistoricalDocumentSynthesis2019a}.
Our template document images are generated based on the specifications from \LaTeX~files; they define the layout, font, size, and content (see Figure~\ref{fig:intro_fig}).
As a text, we use the \textit{Bellum Gallicum}~\cite{edwards1917caesar} with a one or two-column layout.
Additionally, we populated each document with different decorative starting letters.
The advantage of this technique is that we have very precise ground truth, which is the transcription of the document and the exact position of the word on the page.
This dataset contains 455 document images with a resolution of $2754 \times 3564$.

\subsection{Target Domain Dataset}\label{sec:target_dataset}
The target domain dataset refers to the collection of historical documents whose style we aim to learn in the transformation function. 
To create this dataset, we use the historical documents present in the \ac{HBA} 1.0 dataset~\cite{mehriHBAPixelbasedAnnotated2017}. 
The HBA dataset comprises 11 books, where 5 are manuscripts and 6 books are printed, containing 4436 scanned historical document images. 
These books were published between the 13th and 19th centuries and are written in different scripts and languages.
We use one book of this dataset; the handwritten Latin book ``Plutarchus, Vitae illustrium virorum''.
This book contains 730 colored pages with a resolution of $6158 \times 4267$ (see Figure~\ref{fig:dataset_hba}) from which we filtered out 120 pages (blank, binding, and title pages), leaving us with 600 pages that contain only text.
To validate the best model for our downstream evaluation task, we hand-labeled 350 individual word crops from this book.

\subsection{Evaluation Dataset}\label{sec:eval_dataset}
As part of the evaluation process, we use two different datasets.
Our evaluation protocol involves pre-training a \ac{HTR} model using synthetic data generated using our method, and then evaluating it in a fine-tuning setting on the St. Gall dataset~\cite{fischer2011TranscriptionAlignmentLatin} (see Figure~\ref{fig:dataset_sg}).
The Saint Gall dataset includes 40 pages of labeled historical handwritten manuscripts containing 11'597 words and 4'890 word labels.
Each image has a resolution of $3328 \times 4992$ with quality of 300dpi.

To compare our synthetic data pre-training against pre-training on a real handwritten dataset, we pre-train an \ac{HTR} model on the IAM Handwriting Database~\cite{marti2002iam} (see Figure~\ref{fig:dataset_iam}).
This \ac{HTR} model (pre-trained on the IAM Handwriting Database) is then evaluated similarly in a fine-tuning setting on the St. Gall dataset. 
The IAM Handwriting Database contains 1'539 handwritten scanned pages with 115'320 words and 6'625 unique words.
The word images are normalized and in black-white colorization.

\section{Method}\label{ch:MB} 
Our method uses a CycleGAN formulation, along with \ac{HTR} models, to further constrain the synthesis process. 
To train the CycleGAN, we use unpaired collections of user-specified template images (source domain) and real historical images (target domain). 
The source domain documents specify the content and overall structure, and the target domain documents exemplify the style we want in our final synthetic historical documents.

Pondenkandath et al.~\cite{pondenkandathHistoricalDocumentSynthesis2019a} have shown that using only the CycleGAN formulation with the source and target domain datasets is enough to produce synthetic documents that appear stylistically similar to the target domain.
However, they do not contain the content or structure specified in the source domain documents. 
We add a loss term using \ac{HTR} models that aim to read user-specified content from the synthesized historical documents to address this issue.
After completing training, we obtain a generator that transforms any given template image to a corresponding synthetic historical version.

\newpage
\subsection{Model Architecture}\label{sec:model_architecture}
%

Our model architecture is based on the CycleGAN formulation. 
It uses the cycle-consistency loss to transform an image from a given source domain to a target domain in a bi-directional fashion.
This architecture introduces two main challenges.
First, generating text in the target domain that is human-readable at the character and word levels is difficult due to the under-constrained nature of the CycleGAN architecture for our task.
Second, CycleGANs are prone to emergent \ac{GAN} steganography~\cite{zhangSteganoGANHighCapacity2019}; where the generators in a \ac{GAN} can learn to hide information from the discriminator within the synthesized image and use it for perfect reconstruction.

\begin{figure}[t]
    \center{\includegraphics[width=\textwidth]
    {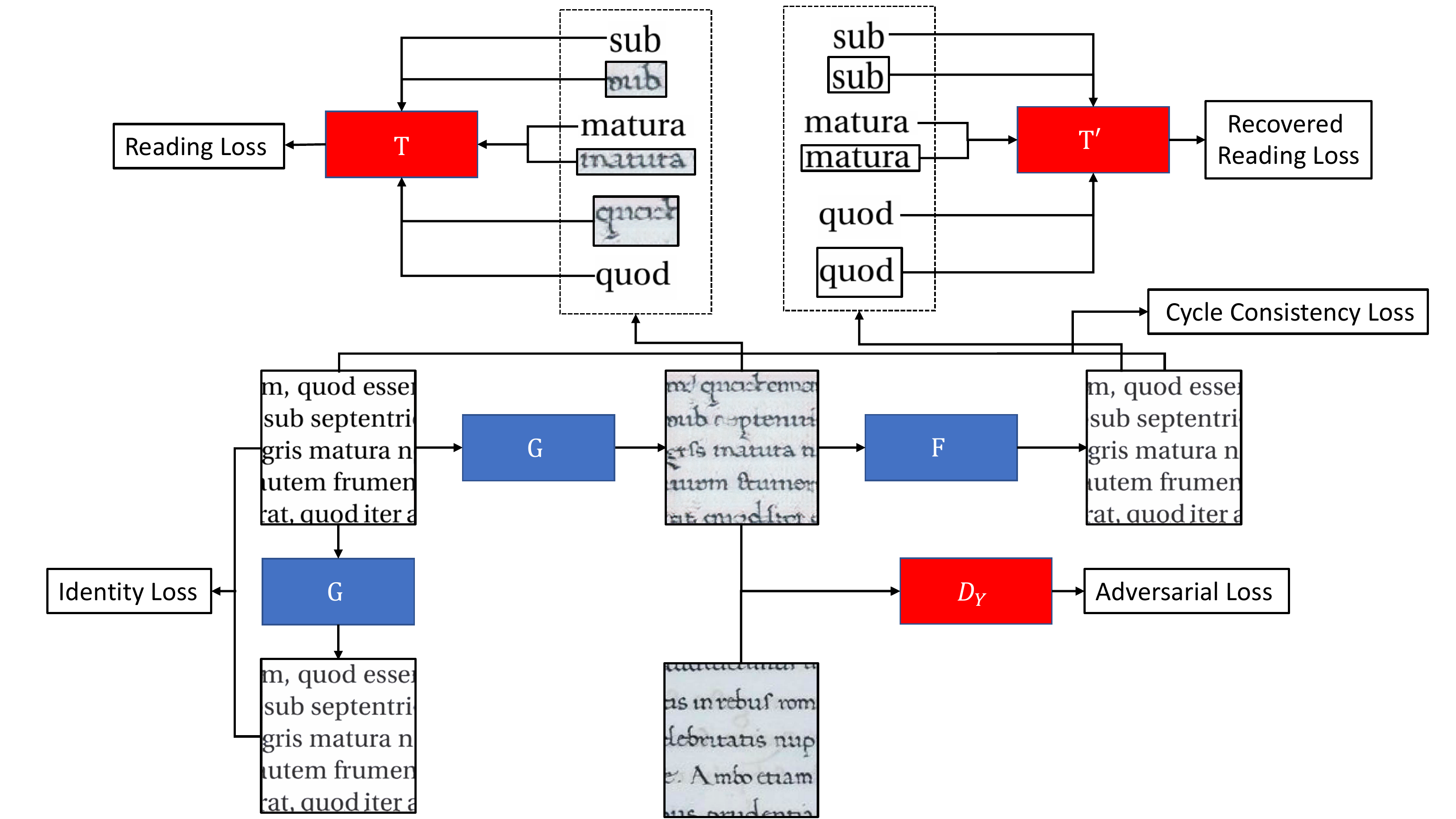}}
    \caption{
    The CycleGAN architecture presented in \cite{zhuUnpairedImagetoimageTranslation2017} with two additional \ac{TR}s $T \text{ and } T^\prime$ and the five different loss terms.
    }
    \label{fig:cycleGAN_architecture}
\end{figure}

To tackle the first problem of generating human-readable text, we introduce two \ac{HTR} models $T \text{ and } T^\prime$ to our architecture (see Figure~\ref{fig:cycleGAN_architecture}).
We aim to adjust for the under-constrained nature of the CycleGAN by adding additional loss terms based on this \ac{HTR} model.
We adopt the bi-directional \ac{LSTM} and \ac{CTC} based \ac{HTR} architecture used by the winners of the text recognition competition at ICFHR'18~\cite{straussICFHR2018CompetitionAutomated2018}.

%
%
The first \ac{HTR} model $T$ evaluates the quality of the characters or words produced by transforming a source domain template image to the target historical domain. 
To do this, it takes as input the synthetic images produced by the source-to-target generator $G$ as well as the textual content and location information from the template document images. 
%
The second \ac{HTR} model $T^\prime$ evaluates the quality of the reconstructed source domain documents (produced by the target-source generator $F$) by comparing the reconstructed image against the same textual content and location information as $T$. 

The second challenge is overcoming the tendency of CycleGANs to hide information about the input within the generated synthetic image~\cite{chuCycleGANMasterSteganography2017}. 
This tendency arises naturally due to cooperation between the generators and is potentially exacerbated by the presence of the \ac{HTR} models.
To minimize the cyclic consistency loss as well as the loss introduced by the \acp{TR}, the generator $G$ attempts to hide information that can be effectively decoded by generator $F$ to produce good reconstructions, as well as information that allows the \acp{TR} to recover the textual content. 
These results in synthetic documents that do not satisfy the constraints of our synthesis process, yet produce very low reconstruction losses and \ac{HTR} losses. 
In some of our preliminary experiments, the generator places the encoded template document into the target document by adding or subtracting the encoded value from each pixel.
The influence on the image is so small that it is nearly impossible for humans to detect, and it is even challenging to be detected by the style discriminator.
Allowing the CycleGAN to cheat prevents it from learning the correct mapping from the target domain back to the source domain, negatively affecting the style representation learned by the \ac{GAN}.
    %

To prevent the CycleGAN from creating this hidden embedding, we add Gaussian noise to the synthetic document images.
This low-frequency noise disturbs the encoded message of the generator, making it much harder to cheat by using steganography.
This noise effectively prevents the network from cheating, as a much stronger signal would be needed, which would manipulate the appearance of the image in a way that is more easily detected by the human eye as well as the style discriminator, and thus would achieve a much lower performance score.

\subsection{Loss Functions}
We train with a loss objective that consists of five different loss terms (see Figure~\ref{fig:cycleGAN_architecture}).
The identity loss, the adversarial loss, and the cycle consistency loss are the loss terms presented in the original CycleGAN paper~\cite{zhuUnpairedImagetoimageTranslation2017}.
To solve the readability problem described in Section~\ref{sec:model_architecture}, we introduce two additional loss terms using the \acp{HTR} system, the reading loss, and the recovered reading loss.
The identity, adversarial, and cycle consistency loss are calculated in both directions, but the reading loss terms are just calculated once per cycle.

Formally, we aim to learn mappings between two domains $X$ and $Y$ with the help of $N$ training samples $x \in X$ and $M$ samples $y \in Y$. 
Each document image $x$ is composed of pairs of its word images and the corresponding word text (ground truth) $x = {((x_{1}, z_{1}), (x_{2}, z_{2}), ..., (x_{n}, z_{n}))}$ where $n = |x|$ and $|x|$ is the amount of words in a document.

The data distributions are denoted as $x \sim p_{data}(x)$ and $y \sim p_{data}(y)$.
We also define a projection function $\alpha$ where $\alpha_1(x)$ refers to the first and $\alpha_2(x)$ to the second element of the tuple. 
The transformation functions of generators $G$ (source-target) and $F$ (target-source) are denoted respectively by $g : X \rightarrow Y$ and $f : Y \rightarrow X$. 
Additionally, we have two adversarial discriminators $D_x$ and $D_y$. 
The task of $D_x$ is to distinguish the images of $\{x\}$ and $\{f(y)\}$, and in the same fashion $D_y$ learns to differentiate between $\{y\}$ and $\{g(x)\}$.

\subsubsection{Identity Loss}

This loss term~\cite{zhuUnpairedImagetoimageTranslation2017,taigmanUnsupervisedCrossDomainImage2016} is used to regularize both generators to function as identity mapping functions when provided with real samples of their respective output domains. 
Zhu et al.~\cite{zhuUnpairedImagetoimageTranslation2017} observed that in the absence of this identity loss term, the generators $G$ and $F$ were free to change the tint between the source and target domains even without any need to do it.
The identity loss is defined as follows:
\begin{align}
    \mathcal{L}_{\text{identity}}(G, F) =& \mathbb{E}_{x\sim p_{\text{data}}(x)}[\norm{G(\alpha_1(x))-\alpha_1(x)}_1] \nonumber \\ 
    +& \mathbb{E}_{y\sim p_{\text{data}}(y)}[\norm{F(y)-y}_1].\lbleq{identity}
\end{align}

\subsubsection{Adversarial Loss}
The adversarial loss~\cite{goodfellowGenerativeAdversarialNetworks2014} shows how well the mapping function $g$ can create images $g(x)$ which looks similar to images in the domain $Y$, while the discriminator $D_y$ aims to tries to distinguish between images from $g(x)$ and real samples from $Y$.
$g$ tries to minimize this objective against $D_y$, which tries to maximize it, i.e. $min_g max_{D_y} \mathcal{L}_{\text{GAN}}(g,D_Y,X,Y)$.
As we use a CycleGAN, this loss is applied twice, once for $g$ and its discriminator $D_y$, as well as for $f$ and the discriminator $D_x$.
\begin{align}
    \mathcal{L}_{\text{GAN}}(g,D_Y,X,Y) =& \mathbb{E}_{y \sim p_{\text{data}}(y)}[\log D_Y(y)] \nonumber \\
   +& \mathbb{E}_{x \sim p_{\text{data}}(x)}[\log (1-D_Y(g(\alpha_1(x)))].\lbleq{GAN}
\end{align}

\subsubsection{Cycle Consistency Loss}\label{sec:ccl}
The cycle consistency loss~\cite{zhuUnpairedImagetoimageTranslation2017} further restricts the freedom of the \ac{GAN}. 
Without it, there is no guarantee that a learned mapping function correctly maps an individual $x$ to the desired $y$. 
Hence, for each pair $(x_\text{i}, z_\text{i}) \in x$ the cycleGAN should be able to bring the image $x_\text{i}$ back into the original domain $X$, i.e. $x_\text{i} \rightarrow g(x_\text{i}) \rightarrow f(g(x_\text{i})) \approx x_\text{i}$.
As the nature of the cycleGAN is bidirectional the reverse mapping must also be fulfilled, i.e. $y \rightarrow f(y) \rightarrow g(f(y)) \approx y_i$.
\begin{align}
    \mathcal{L}_{\text{cyc}}(g, f) =  & \mathbb{E}_{x\sim p_{\text{data}}(x)}[\norm{f(g(\alpha_1(x)))-\alpha_1(x)}_1] \nonumber \\ 
    + & \mathbb{E}_{y\sim p_{\text{data}}(y)}[\norm{g(f(y))-y}_1].\lbleq{cycle}
\end{align}

\subsubsection{Reading Loss and Recovered Reading Loss}
As described in Section~\ref{sec:model_architecture} and shown in Figure~\ref{fig:cycleGAN_architecture}, we use the reading loss to ensure that the \ac{GAN} produces readable images, i.e. images containing valid Latin characters. 
The \acp{TR} $T \text{ and } T^\prime$ are trained with a \ac{CTC} loss~\cite{gravesConnectionistTemporalClassification2006,liReinterpretingCTCTraining2020}, which is well suited to tasks that entail challenging sequences alignments.
\begin{align}
  \mathcal{L}_{CTC}(\mathbf{x, y})=-{\rm ln}\,p(\mathbf{x | y}).
  \label{equ:ctcLoss}
\end{align}
To calculate the reading loss, the template word text $z_\text{i}$ and the corresponding transformed word image $G(x_\text{i})$ is passed to the \acp{TR} $T \text{ and } T^\prime$.
The loss evaluates the mapping $g$ to our target domain $Y$ at a character level.


This discriminator evaluates the readability of the reconstructed image.
Hence, its input is a word text from the source domain $z_\text{i}$ and the respective reconstruction $f(g(x_\text{i}))$.
As above, we calculate the \ac{CTC}-loss on a word level $x_{\text{i}}$. Since the documents all have a different length, the per word losses for each document are summed up and divided by the length of the document $|x|$.

The two reading loss terms are combined to form the overall reading loss defined as
\begin{align}
  \mathcal{L}_{\text{reading}}(g, f) = \mathbb{E}_{x\sim p_{\text{data}}(x)} 
  & \left [\frac{\sum_{v, w \in s(g, x)} \mathcal{L}_{CTC}(\alpha_2(v), w)}{|x|}\right] \nonumber \\
  +& \left[\frac{\sum_{v, w \in s(f(g, x))} \mathcal{L}_{CTC}(\alpha_2(v), w)}{|x|}\right]
  \label{equ:reading_loss}
\end{align}
where $s(h, u) = \{ (u_i, h(u_i))  |  i = 1, ..., |u| \}$ and $h$ represents the transformation function and $u$ all word image and ground truth pair of a document.

\subsubsection{Combined Loss}

The different loss term are weighted with $\lambda_{\text{cyc}} = 10$, $\lambda_{\text{read}} = 1$, and $\lambda_{\text{id}} = 5$ as suggested by Zhu et al.~\cite{zhuUnpairedImagetoimageTranslation2017} and Touvron et al.\cite{touvronPowersLayersImagetoimage2020} and summed up to form the overall loss objective:
\begin{align}
     \mathcal{L_{\text{total}}}(g,f,D_X,D_Y) = & \mathcal{L}_{\text{GAN}}(g,D_Y,X,Y) \nonumber 
    +\ \mathcal{L}_{\text{GAN}}(f,D_X,Y,X) \nonumber \\
    +&\ \lambda_{cyc} \times \mathcal{L}_{\text{cyc}}(g, f) \nonumber
    +\ \lambda_{id} \times \mathcal{L}_{\text{Identity}}(g, f) \nonumber \\
    +&\ \lambda_{read} \times \mathcal{L}_{\text{reading}}(g, f).
    \lbleq{full_objective}
\end{align}
The combined loss is used in a min-max fashion, the generator tries to minimize it, and the discriminators aim to maximize it:
\begin{equation}
    g^*,f^* = \arg\min_{g,f}\max_{D_x,D_Y} \mathcal{L_{\text{total}}}(g, f, D_X, D_Y).
    \lbleq{minmax}
\end{equation}

\section{Experimental Setup}

\subsubsection{Model Architecture}
To achieve the goal of learning a transformation from source domain $X$ to target domain $Y$ using unpaired collections of images, we use an architecture based on the CycleGAN~\cite{zhuUnpairedImagetoimageTranslation2017} framework.
The generators $G$ and $F$ are each 24 layers deep \ac{CNN} architectures with $11.3$ million parameters.
The discriminators $D_x$ and $D_y$ are based on the PatchGAN architecture~\cite{isolaImagetoimageTranslationConditional2017}, and have 5 layers and $2.7$ million parameters each. 
Our \acf{TR} networks $T$ and $T^\prime$ are based on the winning \ac{HTR} model from the ICFHR2018 competition~\cite{straussICFHR2018CompetitionAutomated2018}.
Both these networks contain 10 convolutional and batch normalization layers followed by 2 bi-directional \ac{LSTM} layers for a total of $8.3$ million parameters. 
For all architectures, we apply the preprocessing steps (e.g. resizing) as suggested in their respective publications.
The data gets min-max normalized.

\subsubsection{Task}
The first step in our two-stage method is to create the source domain dataset images as described in Section~\ref{sec:source_dataset}.
The structure and content of these documents are specified using LaTeX.
In the second step, we use the source domain dataset files along with a collection of unlabeled historical document images (see Section~\ref{sec:target_dataset}) to train our CycleGAN and \ac{TR} networks. 
In the training process, we learn a mapping function $g$ that transforms source domain documents to the target domain as well as a mapping function $f$, which works in the other direction.
The \ac{TR} networks are trained simultaneously to recover the user-specified content from $g(x)$ and $f(g(x))$. 
After completing training, we use the generator $G$ to transform document images from the source domain to the target domain while preserving content and structure.

\subsubsection{Pre-processing}
Due to GPU memory constraints, we use to train our models using image patches of size $256 \times 256$. 
These image patches are randomly cropped from the document images and fed into the CycleGAN architecture.
The \ac{TR} networks $T$ and $T^\prime$ receive individual words cropped ($128\times32$) from $g(x)$ and $f(g(x))$ respectively. 
Additionally we add Gaussian Noise to $g(x)$ as described in Section~\ref{sec:model_architecture}.

\subsubsection{Training Procedure}
We train the CycleGAN and \ac{TR} components of our system simultaneously. 
The models are trained for 200 epochs using the Adam optimizer~\cite{kingma2017AdamMethodStochastic} with a learning rate of $2\times 10^{-4}$ and a linear decay starting at 100 epochs. 
The optimizer uses $5\times10^{-5}$ weight decay and $0.5$, $0.999$ beta values for the generators and discriminators, respectively.  
We use a batch size of 1 to facilitate the varying amount of words per patch that is fed to $T$ and $T^\prime$.



\subsubsection{Evaluation Procedure}\label{sec:experimental_setup_evaluation_procedure}
We evaluate the quality of the synthetic historical documents produced with our method qualitatively and quantitatively. 
We first evaluate the synthetic historical documents produced qualitatively with a visual inspection, highlighting the successfully transformed and key limitations of the produced synthetic documents. 

We use synthetic historical documents produced with our method in a pre-training setting to provide a quantitative evaluation. 
We generate 70'000 synthetic words in the historical style of the target domain dataset and use these words to train a new \ac{TR} network called $\mathcal{R}_{\text{syn}}$. 
We then fine-tune $\mathcal{R}_{\text{syn}}$ using various subsets ($10\%$, $20\%$, $50\%$, and $100\%$) of the training data from the St. Gall dataset (see Section~\ref{sec:eval_dataset}) and evaluate its text recognition performance on the test set. 
As baselines, we compare $\mathcal{R}_{\text{syn}}$ against $\mathcal{R}_{\text{base}}$ and $\mathcal{R}_{\text{IAM}}$.
$\mathcal{R}_{\text{base}}$ is randomly initialized and then trained directly on the St. Gall dataset in a similar manner as $\mathcal{R}_{\text{syn}}$.
$\mathcal{R}_{\text{IAM}}$ is pre-trained on the IAM Handwriting Database (see Section~\ref{sec:eval_dataset}) and fine-tuned on the St. Gall dataset. 

To determine the best performing pre-trained models of $\mathcal{R}_{\text{syn}}$ and  $\mathcal{R}_{\text{IAM}}$, we train both networks until convergence and select the best performing model based on validation score from the hand-labeled subset of HBA (see Figure~\ref{fig:hba_words} and validation split of the IAM Handwriting Database. 
The performance of these three models is compared on the test split of the St. Gall dataset using the \acf{CER} and \ac{WER} metrics~\cite{margnerToolsMetricsDocument2014}.

%

\begin{figure}[t]
    \centering
\subfloat[]{{%
  \includegraphics[width=0.15\textwidth]{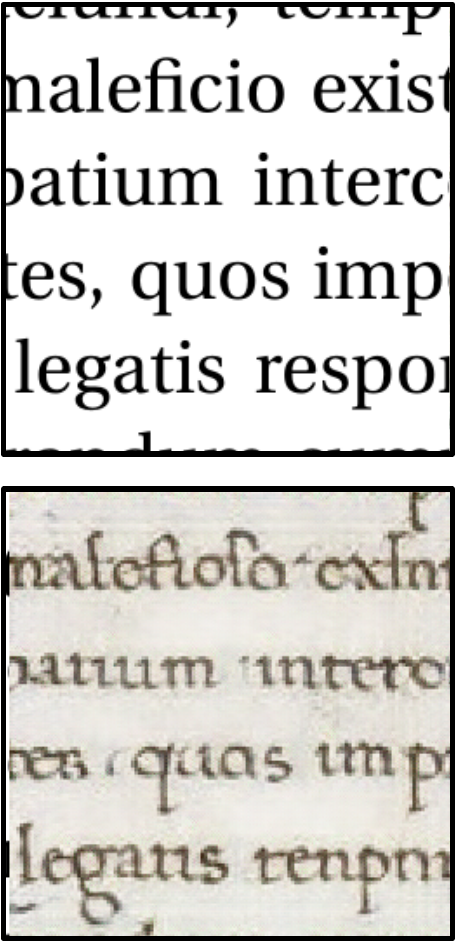}}%
  \label{fig:results_visual-1}}%
\hspace{0.1em}
\subfloat[]{{%
  \includegraphics[width=0.15\textwidth]{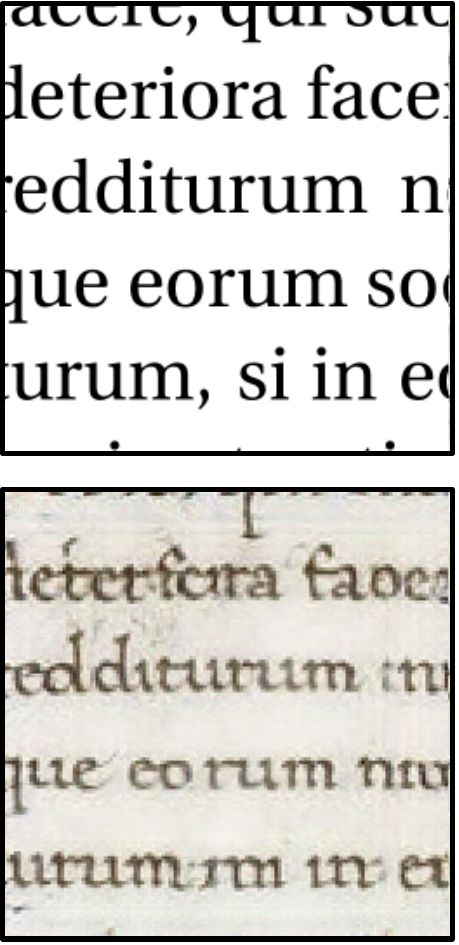}}%
  \label{fig:results_visual-2}}%
\hspace{0.1em}
\subfloat[]{{%
  \includegraphics[width=0.15\textwidth]{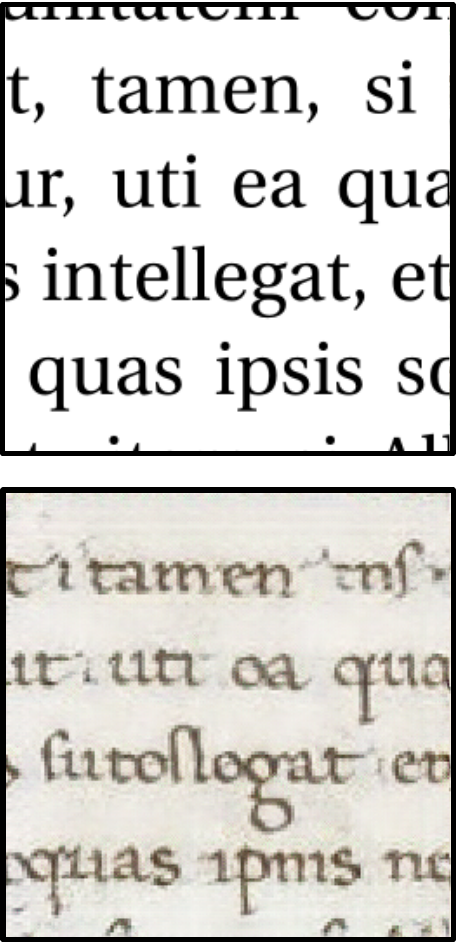}}%
  \label{fig:results_visual-3}}%
\hspace{0.1em}
\subfloat[]{{%
  \includegraphics[width=0.15\textwidth]{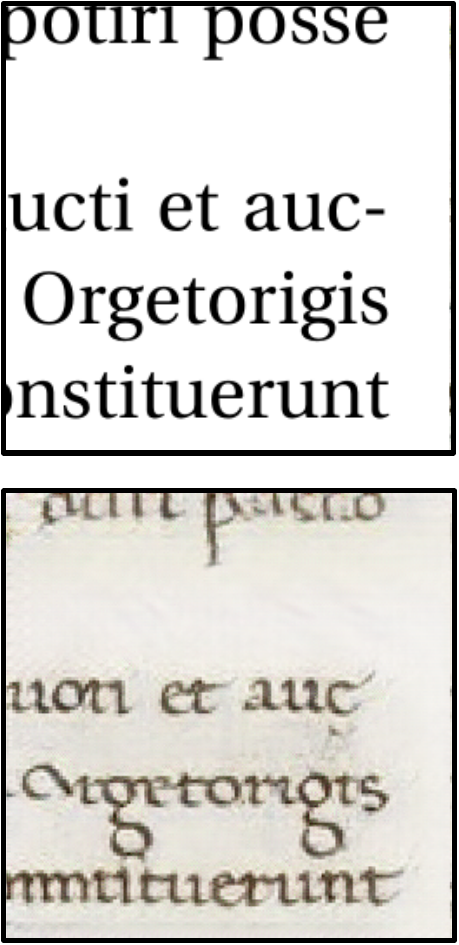}}%
  \label{fig:results_visual-4}}%
\hspace{0.1em}
\subfloat[]{{%
  \includegraphics[width=0.15\textwidth]{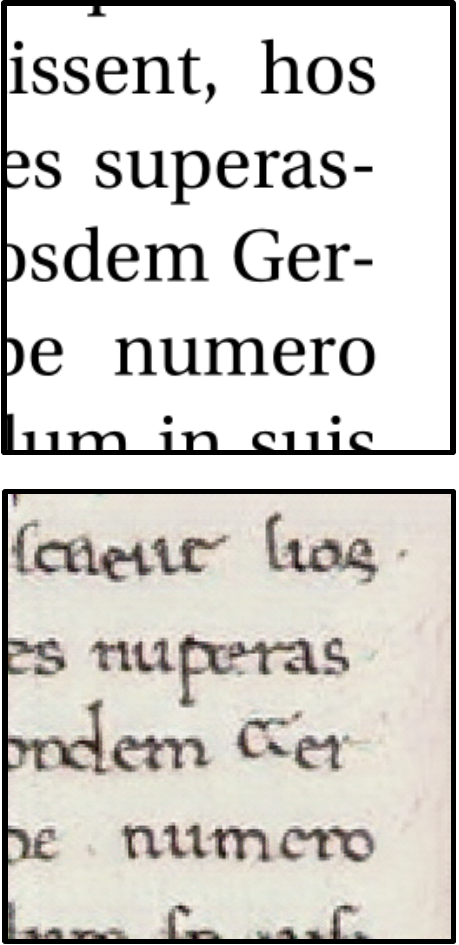}}%
  \label{fig:results_visual-5}}%
\hspace{0.1em}
\subfloat[]{{%
  \includegraphics[width=0.15\textwidth]{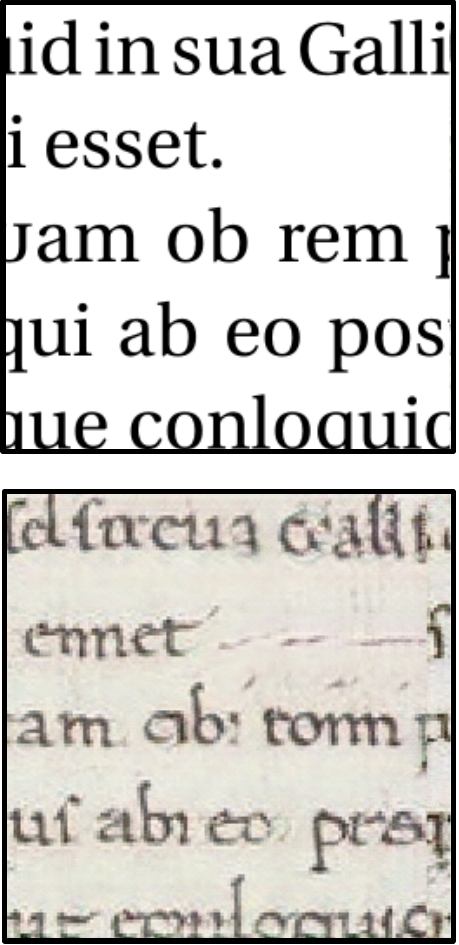}}%
  \label{fig:results_visual-6}}%
    \caption{
   Examples of template documents (upper row) and their corresponding synthetic historical images (bottom row). 
   We can see that most characters are accurately transformed to the historical style while remaining readable. Notable exceptions include `o' $\rightarrow$ `a' and `s' $\rightarrow$ `n'. 
    }
    \label{fig:results_visual}
\end{figure}

\section{Results}
We use two ways to evaluate the results of our generative model: a qualitative visual inspection and a qualitative evaluation.
We use a qualitative human-based approach to evaluate the output from a visual perspective and a qualitative approach to measure the influence of our generated data on a downstream text recognition task.

\subsection{Visual Analysis}

As we can see in Figure~\ref{fig:results_visual}, the synthetic historical documents generated using our method achieve a high degree of similarity to documents from the target domain (see Section~\ref{sec:target_dataset}. 
The two primary goals of our approach were to preserve structure and content during the transformation of the source domain document into the target domain. 

From Figure~\ref{fig:intro_fig}, we can observe that the generator preserves the location of the text from the source domain to the target domain, resulting in the overall structure in the synthetic document matching the input document structure. 
In most cases, the transformation preserves the number of characters, words, and lines from the source document.
However, we observe that on rare occasions, our approach results in synthetic documents where two letters in the source document are combined into a single letter (\textit{legatis} in Figure~\ref{fig:results_visual-1}) or a single letter is expanded into multiple letters (\textit{rem} in Figure~\ref{fig:results_visual-6}).
We can also see from Figure~\ref{fig:intro_synthetic} that our approach is not very effective at transforming the large decorative characters at the beginning of paragraphs. 
The color of these decorative characters is transformed to the historical style, but they appear slightly distorted. 
This effect can be viewed as a side-effect of our training procedure, which does not emphasize transforming the decorative elements apart from the general style discrimination provided by $D_x$ and $D_y$. 
Additionally, we see artifacts where the patches are stitched together because they are generated individually with 10\% overlap and then combined by averaging.

Considering the preservation of textual content, our approach successfully transforms most individual characters to the style of the target domain dataset. 
Individual words are readable and require some effort to distinguish from real historical image samples -- even to expert eyes. 
However, our approach struggles with the transformation of certain letters. 
From Figure~\ref{fig:results_visual-1}, we can see that the character `o' is mistransformed into an `a'.
However, the shape and appearance of these two letters are very similar and often hard to distinguish. 
Our approach also has problems transforming the letter `s'. 
This character is sometimes transformed into the character `n', for e.g., in the word \textit{superas} in Figure~\ref{fig:results_visual-5}, the first `s' is transformed into `n', however the second `s' is correctly preserved.  
Despite these small mistakes, we can observe that overall the method produces a very faithful transformation of the source document into the target historical style while preserving content and structure.

\subsection{Quantitative Evaluation}

\begin{figure}[t]
\centering

\subfloat[CER]{{%
  \includegraphics[width=0.48\textwidth]{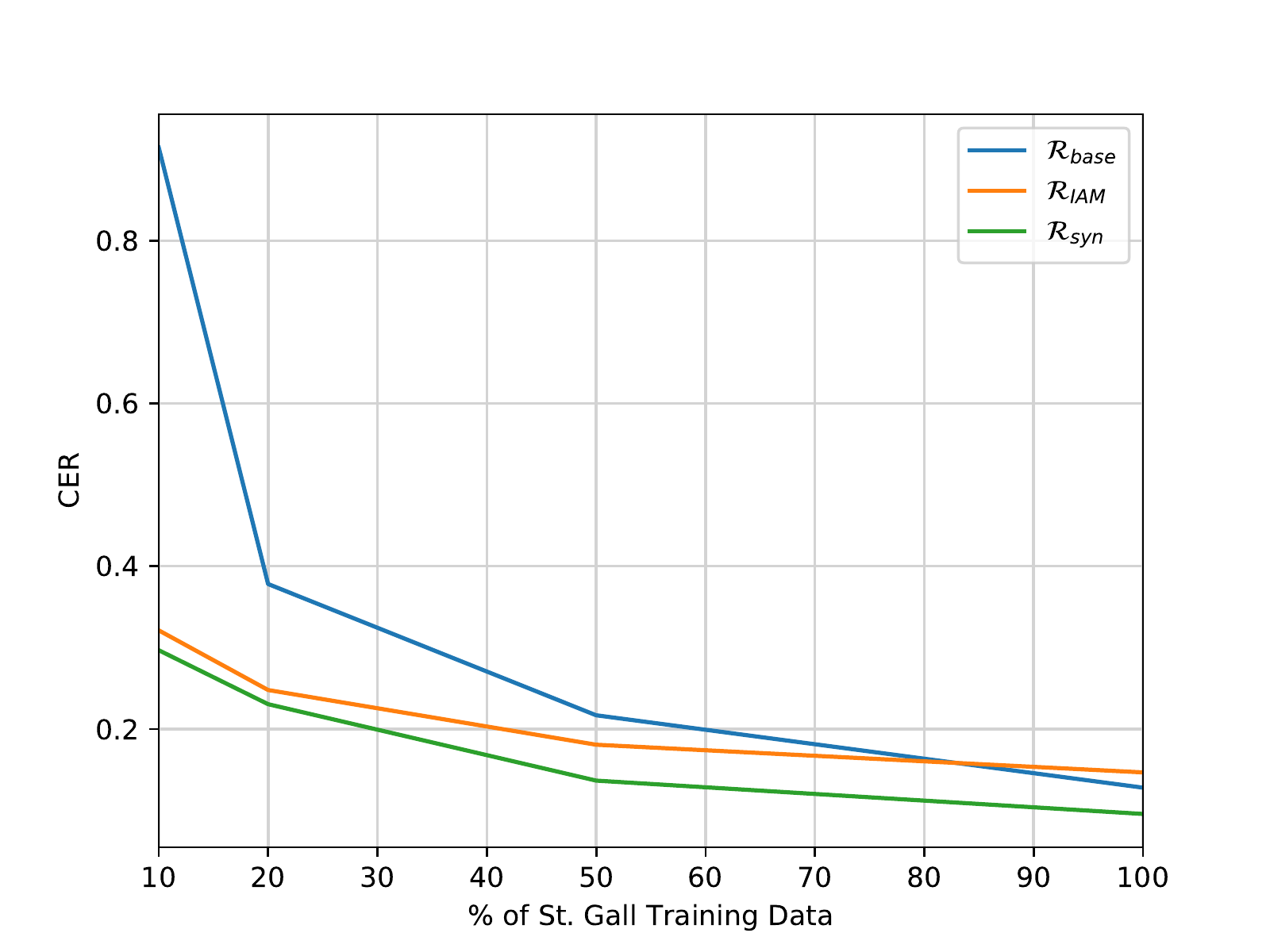}}%
  \label{fig:results_cer_chart}}%
  \quad
\subfloat[WER]{{%
  \includegraphics[width=0.48\textwidth]{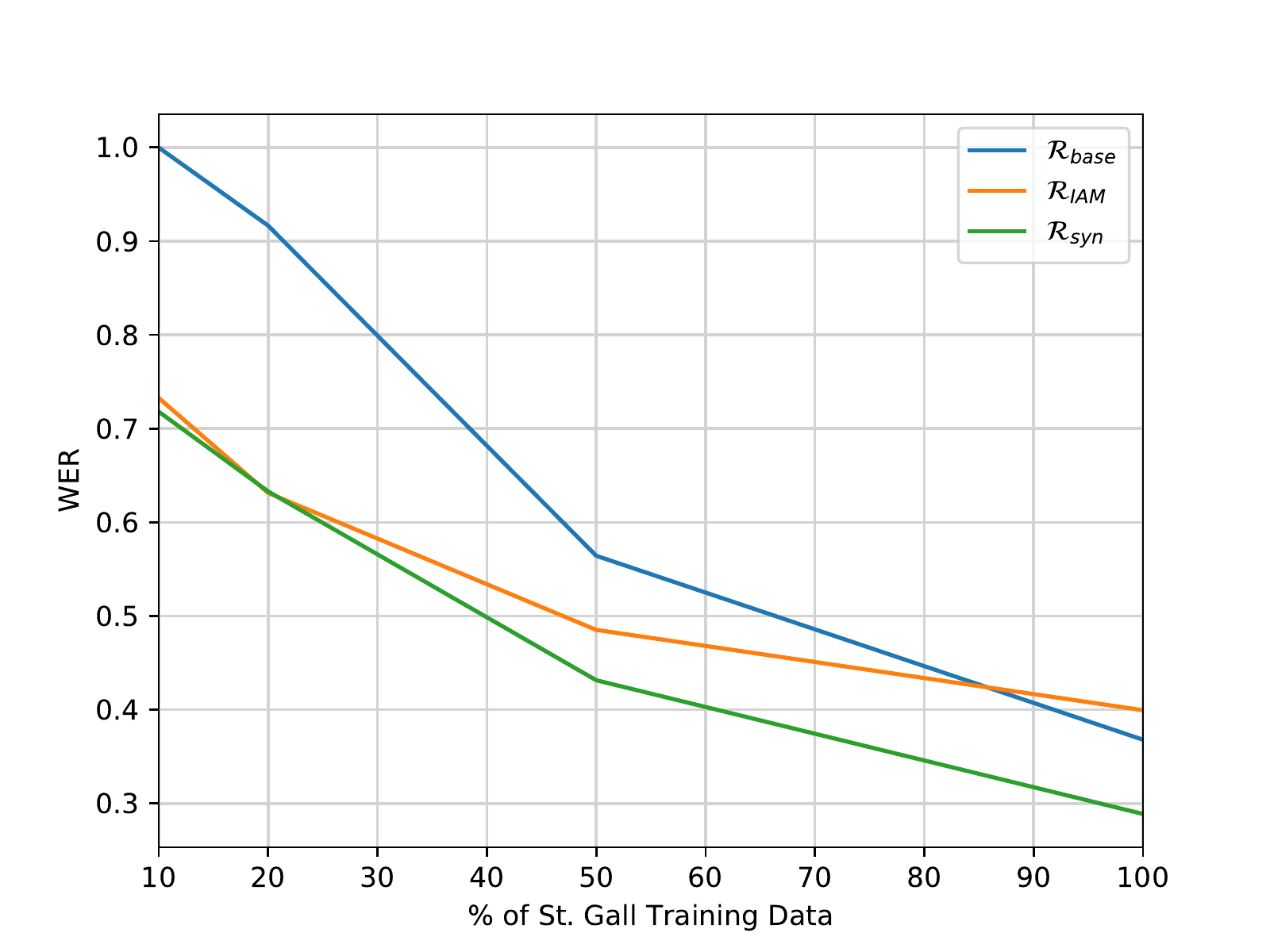}}%
  \label{fig:results_wer_chart}}%
\caption{
We can see that the network pre-trained with synthetic data (in green) outperforms the two baselines (orange and blue) in all categories and for both metrics \ac{CER} and \ac{WER}.}
\label{fig:results_result_plots}
\end{figure}

In Figure~\ref{fig:results_result_plots} we visualize the empirical results of our experiments where we compare our proposed approach against a purely supervised method and a transfer learning baseline method, with respect to the fraction of labels used in the target dataset.
This way, we can assess the performances of those methods in the conditions of arbitrarily (and here, controlled) small datasets. 
We recall that small datasets are the common scenario in this domain, as opposed to more mainstream computer vision domains. As expected, with a small amount of data, the pre-trained methods ($\mathcal{R}_{\text{syn}}$ and $\mathcal{R}_{\text{IAM}}$ ) vastly outperform the baseline ($\mathcal{R}_{\text{base}}$).
This margin decreases as we train on large proportions of training data from St. Gall, however, $\mathcal{R}_{\text{syn}}$ consistently achieves the lowest \ac{CER} (see Figure~\ref{fig:results_cer_chart}), and is narrowly beat by $\mathcal{R}_{\text{IAM}}$ only when considering the \ac{WER} at the $20\%$ subset (see Figure~\ref{fig:results_wer_chart}). 
On average, $\mathcal{R}_{\text{syn}}$ has a $38\%$ lower \ac{CER} and a $26\%$ lower \ac{WER} compared to the model trained only on the St. Gall dataset, and a $14\%$ lower \ac{CER} and $10\%$ lower \ac{WER} compared to the model pre-trained on the IAM Handwriting Database.

Interestingly, when using the entire training set, $\mathcal{R}_{\text{base}}$ achieves a lower error rate than $\mathcal{R}_{\text{IAM}}$, which could be attributed to stylistic differences between the IAM Handwriting Database and the St. Gall dataset. 
Similar to observations from Studer et al.~\cite{studerComprehensiveStudyImageNet2019}, the benefits of pre-training on a different domain could decrease when more training data is available from the actual task. 
Therefore, the stylistic similarity of the synthetic historical images and documents from the St. Gall dataset could explain the lower error rates of $\mathcal{R}_{\text{syn}}$ compared to $\mathcal{R}_{\text{base}}$.


%
%

%

\section{Conclusion}

We presented a two-step framework for generating synthetic historical images that appear realistic.
The two steps are (1) creating electronic user-defined datasets (e.g., with LaTeX) for which the text content is known, and then feed it to step (2) where we use an improved CycleGAN based deep learning model to learn the mapping to a target (real) historical dataset.
Differently from previous works in the field, our approach leverages two \acf{TR} networks to constrain the learning process further to produce images from which the text can still be read.
The outcome of the process is a model capable of synthesizing a user-specified template image into historical-looking images.
The content is known, i.e., we have the perfect ground truth for all the synthetic data we generate.
These synthetic images --- which come with a \ac{OCR} ground truth --- can then be used to pre-train models for downstream tasks.
We measured the performances of a standard deep learning model using images created with our approach as well as other existing real historical datasets. 
We show that our approach consistently outperforms the baselines through a robust set of benchmarks, thus becoming a valid alternative as a source dataset for transfer learning.
This work extends the already conspicuous work on the field of synthetic document generation. 
It distinguishes itself for providing the ground truth and high-quality synthetic historical images. 
Finally, the images generated with our methods are still distinguishable from real genuine ones due to small imperfections. 
Therefore we envisage that further work would improve upon our open-source implementation.

\section*{Acknowledgment}
The work presented in this paper has been partially supported by the HisDoc III project funded by the Swiss National Science Foundation with the grant number 205120\_169618.
A big thanks to our co-workers Paul Maergner and Linda Studer for their support and advice.

%
%
\bibliographystyle{splncs04}
\bibliography{related_work}

\end{document}